\documentclass[11pt]{article}

\usepackage[final]{acl}

\usepackage{times}
\usepackage{latexsym}

\usepackage[T1]{fontenc}
\usepackage{booktabs}
\usepackage{graphicx}
\usepackage{tikz}
\usepackage{blindtext}
\usepackage{cuted}
\usepackage{hyperref}
\usepackage{varwidth}
\usepackage[dvipsnames]{xcolor}
\usepackage[utf8]{inputenc}

\usepackage{microtype}

\usepackage{inconsolata}

\usepackage{graphicx}

\usepackage{amsmath}
\usepackage{amssymb}

\usepackage{twemojis}

\usepackage{siunitx}
\usepackage{enumitem}

\title{
Understanding the Behaviors of Environment-aware Information Retrieval
}

\author{
\textbf{Ruifeng Yuan}$^{1,2}$*\ \ 
        \textbf{Chaohao Yuan}$^{2,3}$*\ \ 
        \textbf{David Dai}$^4$* \ \
        \textbf{Yu Rong}$^{2}$ \ \
        \textbf{Hong Cheng}$^3$ \ \
        \\
        \textbf{Hou Pong Chan}$^2$\footnotemark[2] \ \ 
        \textbf{Chenghao Xiao}$^5$\footnotemark[2] \\\\
  $^1$Fudan University
  $^2$Alibaba DAMO Academy
  $^3$Chinese University of Hong Kong \\
  $^4$Stanford University
  $^5$Shanghai University of Finance and Economics
  \\
}

\begin{document}

\maketitle

\renewcommand{\thefootnote}{\fnsymbol{footnote}}
\footnotetext[1]{Equal Contribution.}
\footnotetext[2]{Corresponding authors, \href{mailto:kenchanhp@gmail.com}{kenchanhp@gmail.com}, \quad\quad\quad \href{mailto:xiaochenghao@sufe.edu.cn}{xiaochenghao@sufe.edu.cn}}

\begin{strip}
    \vspace{-2.5cm}
    \centering
    \begin{minipage}{\textwidth}
        \centering
        \begin{minipage}[t]{0.32\textwidth}
            \centering
            \includegraphics[width=\linewidth, height=5cm, keepaspectratio=false]{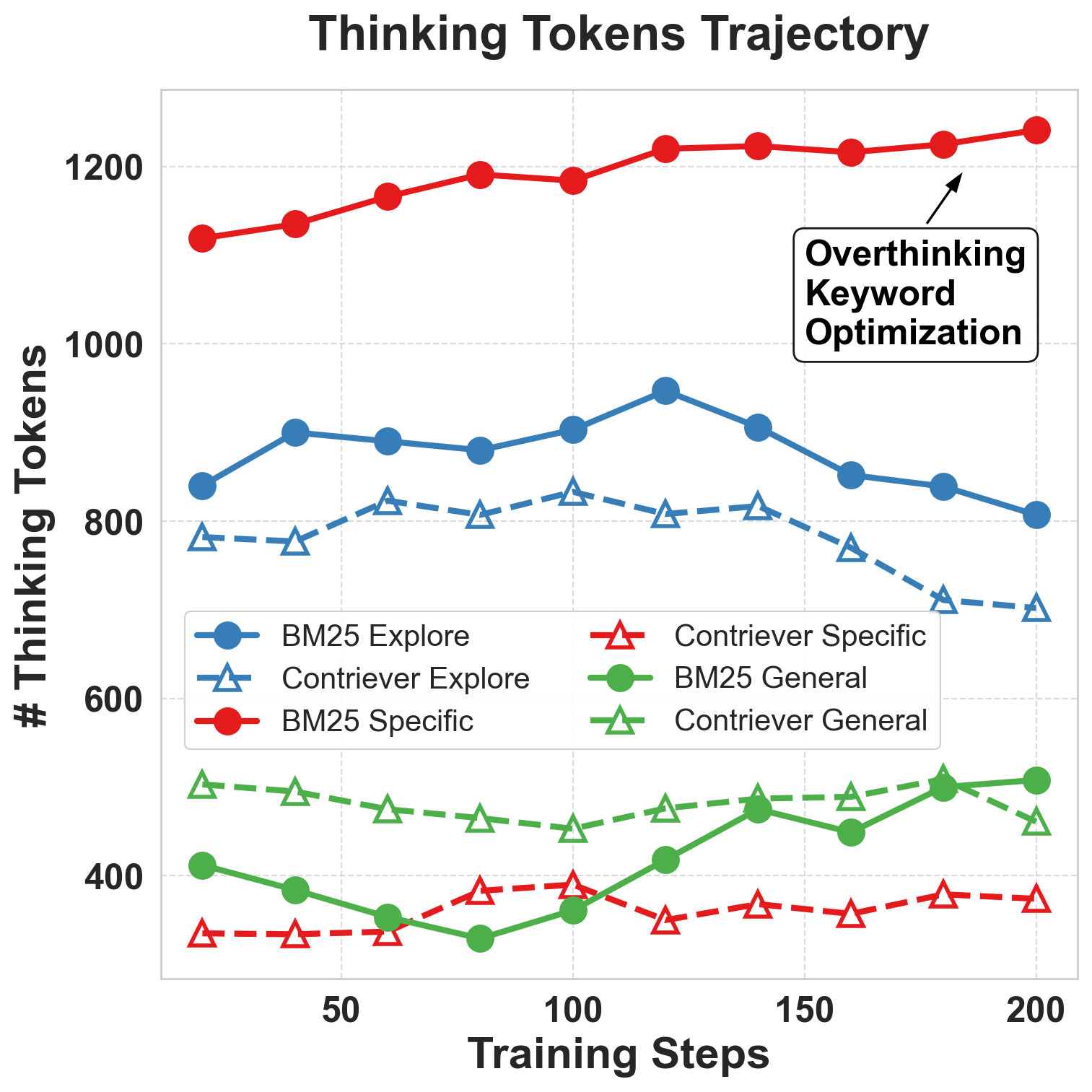} 
            \par \vspace{2pt}
            \small (a) Thinking tokens length dynamics across GRPO steps.
        \end{minipage}
        \hfill
        \begin{minipage}[t]{0.32\textwidth}
            \centering
            \includegraphics[width=\linewidth, height=5cm, keepaspectratio=false]{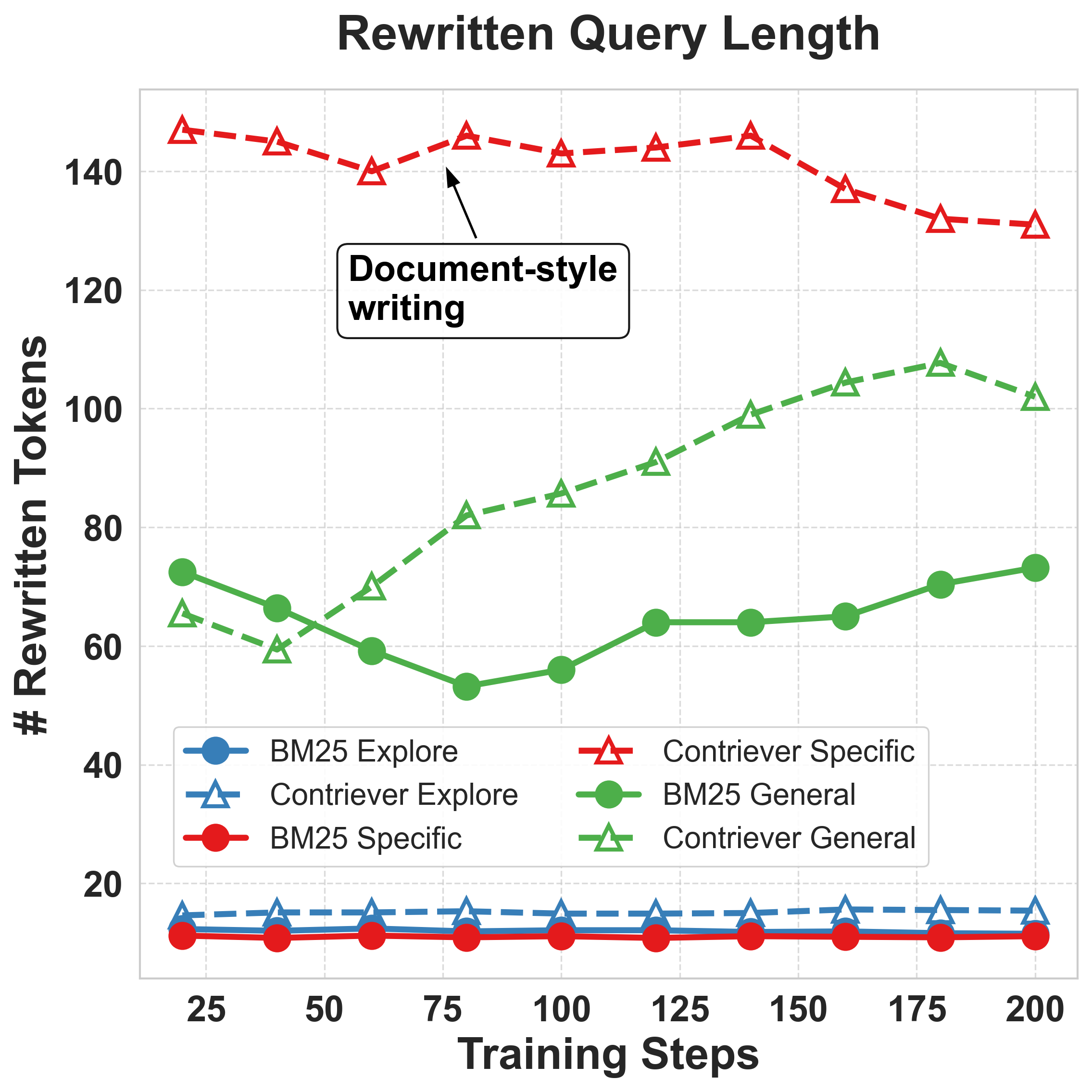}
            \par \vspace{2pt}
            \small (b) Rewritten queries tokens length dynamics across GRPO steps.
        \end{minipage}
        \hfill
        \begin{minipage}[t]{0.32\textwidth}
            \centering
            \includegraphics[width=\linewidth, height=5cm, keepaspectratio=false]{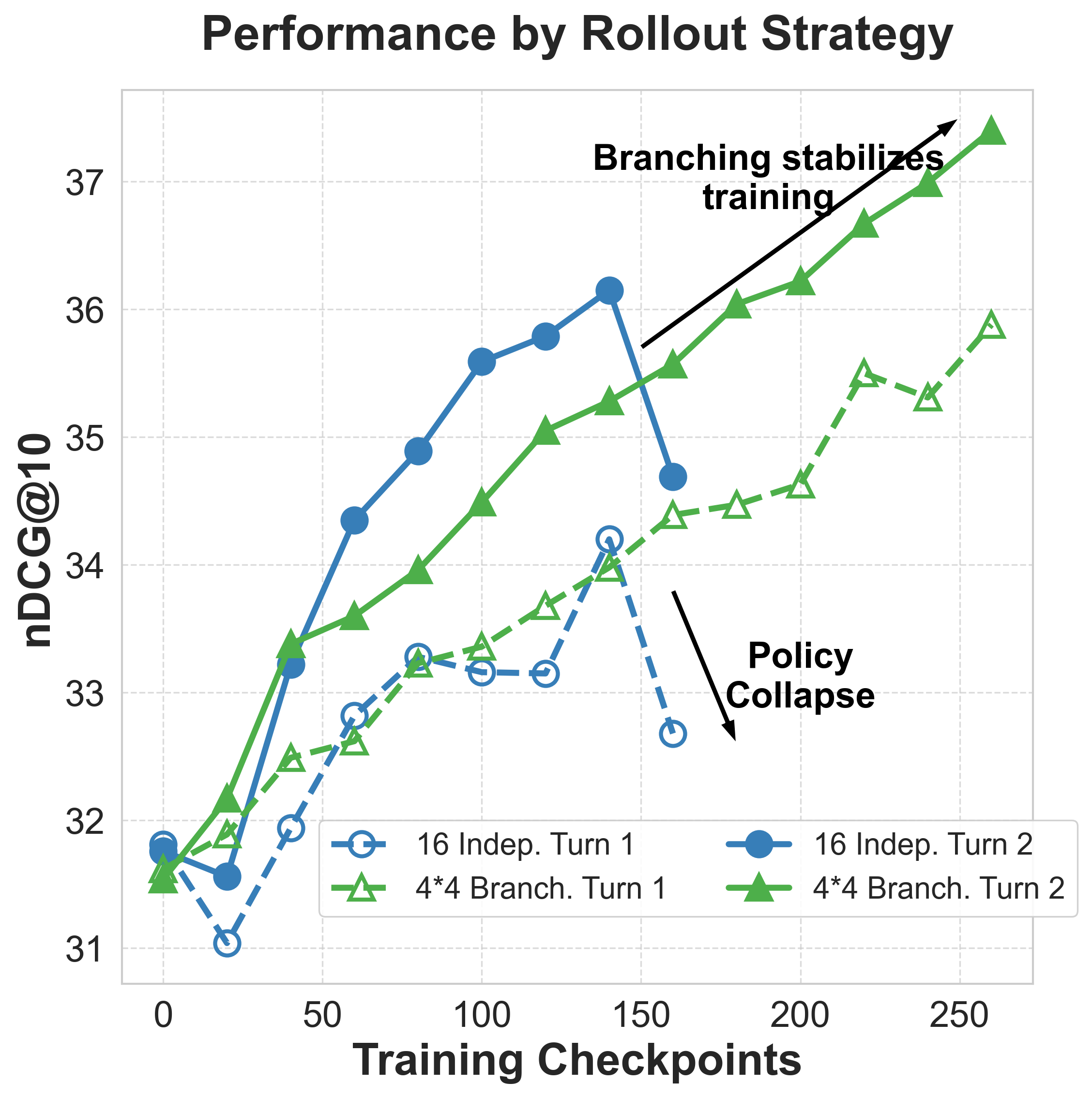}
            \par \vspace{2pt}
            \small (c) Dynamics of two-turn retrieval reward, across rollout techniques.
        \end{minipage}
    \end{minipage}
    
    \par 
    \vspace{0.3cm} 

    \begin{minipage}{\textwidth}
        \centering
        \begin{minipage}[t]{0.32\textwidth}
            \centering
            \fbox{\begin{minipage}[t][6.2cm]{0.92\linewidth}
                \scriptsize 

\textcolor{teal}{\textbf{User Question:} Retrieve a Wikipedia article that answers: Who wrote God's Plan? The retriever you will be using is Contriever.}

\textcolor{purple}{Assistant: \textbf{<think>}}\textcolor{gray}{\{... Omitted...\}} \textcolor{purple}{Since Contriever uses document-to-document retrieval based on spans from Wikipedia, \textbf{I should create a document that includes key information about the song.}} I need to make sure the document includes 
\textcolor{gray}{\{... Omitted...\}}  
Putting it all together, the \textcolor{purple}{\textbf{hypothetical document} should have the song title as the main heading, followed by the artist, release year, genre, writer, and a brief note on its success.} 
\textcolor{purple}{
\textbf{</think>}}
\textcolor{blue}{\textbf{<rewrite>}"God's Plan" was written and produced by Canadian rapper and songwriter Drake. Released in 2018 as the lead single from his fifth studio album "Scorpion", the track features a minimalist production with a repetitive piano loop and a soulful vocal sample. "God's Plan" received widespread acclaim for its introspective lyrics and became a commercial success, reaching number one on the Billboard Hot 100 chart. \textbf{</rewrite>}}
            \end{minipage}}
            \par \vspace{2pt}
            \small (d) Query rewriter learns to rewrite queries as document-style which Contriever favors.
        \end{minipage}
        \hfill
        \begin{minipage}[t]{0.32\textwidth}
            \centering
            \fbox{\begin{minipage}[t][6.2cm]{0.92\linewidth}
                \scriptsize 
            \textcolor{teal}{\textbf{User Question:} who does dwyane wade play for right now? You will be using bm25 to retrieve.}

        \textcolor{purple}{Assistant: \textbf{<think>}}\textcolor{gray}{\{... Omitted...\}} First, I need to figure out the best way to phrase this query for BM25. \textcolor{purple}{\textbf{BM25 is a keyword-based retrieval system}, so including relevant keywords is crucial.} The original query is "who does dwyane wade play for right now." The key elements here are "Dwyane Wade," "play for," and "right now." Since BM25 relies on matching keywords in the document, I should make sure these terms are included. \textcolor{purple}{However, "right now" might be a bit too specific or could be interpreted as a time-related keyword.} Maybe replacing "right now" with "current" or "now" could be better. \textcolor{purple}{Also, using "current team" instead of "play for right now" might be more effective.} 
        \textcolor{gray}{\{... Omitted...\}} 
        Between these, "Dwyane Wade current team" is more likely to yield the correct results because it's a common way to phrase such questions.

        \textcolor{blue}{\textbf{<rewrite>}Dwyane Wade current team 
        \textbf{</rewrite>}}
        
            \end{minipage}}
            \par \vspace{2pt}
            \small (e) For bm25, the query rewriter learns to rewrite queries as keywords-style.
        \end{minipage}
        \hfill
        \begin{minipage}[t]{0.32\textwidth}
            \centering
            \fbox{\begin{minipage}[t][6.2cm]{0.92\linewidth}
                \scriptsize 
\textcolor{teal}{\textbf{User Question:} Jim Price appeared on the Rolling Stones Album released in what month?}

\textcolor{purple}{Assistant:} \textcolor{gray}{\{... Omitted...\}}\textcolor{purple}{<rewrite>}\textcolor{gray}{\{... Omitted...\}}\textcolor{purple}{</rewrite>
}
Retrieved Doc: 
\textcolor{gray}{\{... Omitted...\}}
\textcolor{purple}{Assistant:} \textcolor{gray}{\{... Omitted...\}} But looking at the top documents, doc0 mentions Jim Price touring with The Rolling Stones from 1970-1973 and appearing on "Sticky Fingers", "Exile on Main St.", and "Goats Head Soup". Wait, 
in the initial retrieval, the first query might have assumed "Steel Train" was released in May, but maybe that's not confirmed in all sources. 
\textcolor{gray}{\{... Omitted...\}}\textcolor{blue}{\textbf{<rewrite2>}Jim Price was a horn session player for The Rolling Stones, appearing on their albums "Sticky Fingers," "Exile on Main St.," and "Goats Head Soup" during his tenure from 1970 to 1973. To determine the exact month of release for the Rolling Stones album that featured Jim Price, it is necessary to identify which of these albums were released in a specific month.\textbf{</rewrite2>}}
            \end{minipage}}
            \par \vspace{2pt}
            \small (f) Two-turn environment-aware retrieval.
        \end{minipage}
    \end{minipage}

    \vspace{0.5em}
    \begin{center}
        \begin{minipage}{0.9\textwidth}
            \centering
            \small            
            \textbf{Figure 1: (a-c) GRPO learning Dynamics across retrievers and training strategies. (d-f) Case studies.} 
            \label{fig:teaser}
        \end{minipage}
    \end{center}
    
    \vspace{-0cm} 
\end{strip}
\setcounter{figure}{1} 

\begin{abstract}
Recent retrieval-augmented generation (RAG) approaches have demonstrated strong capability in handling complex queries, yet current research overlooks a critical challenge: different retrievers require fundamentally different query formulation strategies for optimal performance. In this work, we present the first systematic analysis of how LLMs can learn to adapt their query formulation strategies for different retrievers via reinforcement learning (RL). Our empirical study reveals that RL effectively teaches an LLM to tailor its queries to specific retriever characteristics. We discover that different retrievers exhibit surprisingly distinct optimal query styles (e.g., descriptive vs. question-like), suggesting strategies learned for one retriever ineffective for another. We further show that performance can be enhanced by incorporating retriever-specific human guidance and by scaling model size. To facilitate learning over multi-retrieval-step trajectories, we introduce a branching-based rollout technique that improves training stability. Our work provides the first empirical evidence and actionable insights for building truly retriever-aware RAG systems. Code and resources are available at \href{https://github.com/LCO-Embedding/Envs-aware-Information-Retrieval}{LCO-Embedding/Envs-aware-Information-Retrieval}.
\end{abstract}

\section{Introduction}

Retrieval-augmented generation (RAG) is a standard approach for augmenting the knowledge of large language models (LLMs) by generating queries and retrieving relevant information from external sources. Recently, Agentic RAG has emerged as a more sophisticated paradigm that, rather than performing a single retrieval step, decomposes complex queries and iteratively executes multiple retrieval operations to fulfill complex task requirements, such as generating comprehensive research reports in real-world applications.

Most agentic RAG approaches treat retrieval as a uniform ``tool call,'' operating under the assumption that it occurs exclusively through a single commercial search engine (e.g., Google) and optimizing for that specific engine. However, retrieval from locally stored information repositories remains critically important in real-world applications, including private financial reports and similar patient case databases. In practice, agentic RAG systems must accommodate diverse retrieval backends. Different retrievers may require distinct optimal query formulation strategies to maximize retrieval effectiveness—an LLM should fundamentally adapt how it formulates queries depending on which retriever it is querying. However, current research lacks adequate investigation into how LLMs can develop retriever-aware query formulation capabilities.

In this work, we present \textbf{the first systematic analysis of how LLMs learn to adapt their query formulation strategies for different retrievers via reinforcement learning (RL)}. We systematically examine the effects of different retriever types, the comprehensiveness of human guidance in prompts, and the scaling of model size and retrieval steps, via the following research questions. 

\textbf{RQ1: Can LLMs adapt their query formulation strategies to local retrievers via RL? If so, are strategies learned for one retriever transferable to others?
}
We find that RL effectively adapts an LLM's query formulation strategy to local retrievers. Moreover, we discover that different retrievers exhibit distinct optimal query formulation strategies. For example, \texttt{Contriever}~\cite{contriever} requires long-form descriptive queries while \texttt{Qwen3-Embedding}~\cite{zhang2025qwen3} performs better with question-like queries. Consequently, \textit{strategies optimized for one retriever cannot be effectively transferred to different retrievers.}

\textbf{RQ2: How does the comprehensiveness of human guidance in prompts affect the learning of query formulation?}
We design three levels of knowledge comprehensiveness in human guidance: from general query rewriting strategies to retriever-specific query formulation by human experts. We find that more comprehensive retriever-specific knowledge in prompts leads to better LLM performance, suggesting that LLMs benefit from human expertise to guide RL exploration in query formulation.

\textbf{RQ3: Can we enhance query formulation performance by scaling up model size and the number of retrieval steps?}
We find that larger models achieve better performance and can even discover policies that human experts do not anticipate. 
Moreover, to support multi-step retrieval trajectories during rollouts, we introduce a branching technique that isolates the learning signal to the final turn. Experimental results show that increasing the number of retrieval steps enhances query formulation capability, as the model can leverage documents retrieved in earlier steps to refine subsequent query formulation strategies.

We summarize our key contribution as follows: (1) We present the first systematic empirical study analyzing how LLMs learn to adapt their query formulation strategies for different retrievers via reinforcement learning. (2) Our results reveal distinct optimal query formulation strategies for prominent retrievers, providing actionable insights for RAG system practitioners. (3) We introduce a branching-based optimization technique to support multi-step retrieval trajectory rollouts, facilitating stable RL training for multi-step query strategy optimization.

\section{Methodology}
\label{sec:methodology}
We formulate retriever-aware query adaptation as a reinforcement learning (RL) problem, in which an LLM–based query rewriter policy $\pi_\theta$ learns to generate optimal search queries $q$ for a given black-box retriever environment $\mathcal{E}$, conditioned on a user-issued question. The policy is optimized using Group Relative Policy Optimization (GRPO)~\cite{GRPO}, which enables stable learning without requiring a value-function critic.

\subsection{Reward Function Formulation}
A critical component of our framework is the reward signal provided by the environment. For a generated query $q$, the retriever returns a set of documents $D$. We define the reward $r(q, D)$ as the \textbf{Normalized Discounted Cumulative Gain (nDCG@10)} of the retrieved documents $D$ with respect to the ground-truth passage $d^*$. We select nDCG over simple recall to incentivize the model to not only find the correct document but to rank it highly, mirroring real-world RAG requirements.

\subsection{Scenario 1: Single-Turn Retrieval}
In the single-turn setting, the policy generates a rewrite $q$ given an input $x$. For each input, we sample a group of $G$ outputs $\{q_1, ..., q_G\}$ from the policy $\pi_{\theta_{old}}$. The GRPO objective is maximizing:
\begin{equation}
\resizebox{\hsize}{!}{$
\begin{split}
    \mathcal{L}_{GRPO}(\theta) = -\frac{1}{G} \sum_{i=1}^G \frac{1}{|y_i|} \sum_{t=1}^{|y_i|} \bigg[ 
    &\min \left( \rho_{i,t} A_i, \text{clip}(\rho_{i,t}, 1-\epsilon, 1+\epsilon) A_i \right) \\
    &- \beta \mathbb{D}_{KL}(\pi_\theta || \pi_{ref}) \bigg] \text{,}
\end{split}
$} 
\end{equation}

where $\rho_i = \frac{\pi_\theta(q_i|x)}{\pi_{\theta_{old}}(q_i|x)}$ is the importance sampling ratio. The advantage $A_i$ is computed by normalizing the reward $r_i$ against the group statistics:

\begin{equation}
A_i = \frac{r_i - \mu_{group}}{\sigma_{group} + \delta}\text{,}
\label{eq:advantage}
\end{equation}
where $\mu_{group}$ and $\sigma_{group}$ are the mean and standard deviation of rewards within the group.

\subsection{Scenario 2: Multi-Turn Iterative Retrieval}
To enable the agent to refine queries based on initial feedback, we define a multi-turn trajectory $\tau_i = [q_{i,1}, e_{i,1}, q_{i,2}, e_{i,2}, ..., q_{i,K}]$, where $q_{i,k}$ is the query (and reasoning) at turn $k$, and $e_{i,k}$ represents the search results (masked during training).

\paragraph{Weighted Reward Aggregation.} To value the final answer while encouraging useful intermediate steps, we compute the total trajectory reward $R_i$ as a weighted sum:

\begin{equation}
R_i = \sum_{k=1}^{K} \lambda_k \cdot r(q_{i,k}, e_{i,k})\text{.}
\label{eq:multiturn_reward}
\end{equation}

In practice, we assign a higher weight to later rounds, which is discussed in implementation details and multi-round training techniques section.

\subsection{Branching Rollout Strategy}
\label{subsec:branching}
A key challenge in multi-turn RL is the high variance of intermediate rewards, which leads to training collapse. Hence, we introduce a \textbf{Branching Rollout} strategy that stabilizes credit assignment.

In this work, we focus on a two-turn RL setting. Instead of independent linear trajectories (e.g., $16 \times 1$), we implement a branching structure (e.g., $4 \times 4$) where we sample $M$ first-turn actions, and for each, branch into $N$ second-turn completions. This yields two benefits:

\begin{itemize}[leftmargin=*,topsep=2pt]
\setlength{\itemsep}{2pt}
\setlength{\parskip}{0pt}
\setlength{\parsep}{0pt}
    \item \textbf{Monte Carlo Estimation for Turn 1:} The value of a first-turn action $q_{i,1}$ is estimated by averaging over its $N$ branches, providing a robust signal for the intermediate step:
    \begin{equation}
    V(q_{i,1}) \approx r(q_{i,1}) + \frac{1}{N}\sum_{j=1}^{N} r(q_{i,2}^{(j)})\text{.}
    \end{equation}
    
    \item \textbf{Variance Reduction for Turn 2:} By grouping the $N$ branches that share the same history, the baseline (mean reward) becomes specific to that sub-group. This isolates the learning signal for the second turn, ensuring that the advantage calculation reflects the quality of the second rewrite independent of the first turn's quality.
\end{itemize}

\section{Experiments}
\subsection{Settings}

\paragraph{In-distribution Evaluation} 
We leverage RAGBench \cite{friel2024ragbench}, a large-scale benchmark for evaluating retrieval-augmented generation systems. RAGBench comprises more than 100,000 samples, spans a broad range of general and specialized domains, and exhibits substantial variation in document structure and information needs, which makes it well suited for a comprehensive evaluation of model performance.
To adapt RAGBench for retrieval tasks, we first aggregate all documents in the dataset into a global retrieval corpus, and treat the questions as search queries and their associate contexts as the positive pair for retrieval. Additionally, for corpus de-duplication, we employ Qwen-Embedding-0.6B~\cite{zhang2025qwen3} to encode the documents, retaining only one instance from pairs that share a similarity greater than 0.99.

\paragraph{OOD Evaluation} 
To evaluate the transferability of LLM query rewriter trained on general-domain RAG datasets like RAGBench, we conduct zero-shot experiments to the commonly-used BEIR benchmark \cite{thakur2021beir}.

\paragraph{Domain-specific Transfer} To evaluate how well retriever-aware query rewriters trained on general-domain datasets generalize to challenging domain-specific environments, we conduct an in-depth case study in the financial domain. Although underexplored in existing retrieval benchmarks, finance is an ideal stress test for cross-domain transfer due to (1) its pronounced distribution shift from general-domain corpora induced by domain-specific terminology (e.g., SG\&A, currency fluctuation, gross margin), (2) the hybrid structure of financial documents that interleaves text with tabular content, and (3) the practical relevance of information retrieval in real-world workflows in finance. To this end, we construct a financial retrieval environment leveraging FinAgentBench \cite{choi2025finagentbench} and describe the process in detail in Appendix \ref{apx: FinAgentBench}.

\subsection{Implementation Details}

\begin{figure}[t!]
    \centering
    \fbox{
        \begin{varwidth}{0.93\linewidth}
        \small
            \textbf{Question-to-Statement Rewriting Policy discovered by a 14B agent under Contriever environment.} \\
            \vspace{0.2em} 
            \small
\vspace{-2pt}
User Question: Is it possible to be white and latino?

\textcolor{purple}{assistant: <think>}
Okay, let's tackle this question. The user is asking if it's possible to be both white and Latino. I need to figure out the best way to rewrite the query for Contriever to get the most relevant documents.

\textcolor{purple}{({\large \texttwemoji{thinking face}}) First, I should understand what Contriever does.} It's trained on Wikipedia articles, using two spans as positive pairs. So, the model is good at finding documents where two phrases are related. The original query is about the intersection of race and ethnicity, specifically being white and Latino.
\textcolor{gray}{\{... More verbose reasoning omitted\}...}
To get the best results, the rewritten query should capture the essence of the question. \textcolor{purple}{({\large \texttwemoji{light bulb}}) Instead of a direct question, maybe \textbf{a statement that reflects the core of the query.}} For example, "White individuals identifying as Latino" or "Intersection of race and ethnicity in Latino communities." \textcolor{purple}{({\large \texttwemoji{thinking face}}) But I need to make sure it's a query that Contriever can handle effectively.}

Another angle is to use terms that are more likely to appear in Wikipedia articles. Terms like "ethnicity and race intersection," "Latino identity," or "multiracial identity." Maybe combining both terms in a way that Contriever can retrieve documents discussing both aspects.

\textcolor{purple}{({\large \texttwemoji{thinking face}\texttwemoji{light bulb}})Wait, the original question is a yes/no question, but Contriever might perform better with a more descriptive query.} 
Alternatively, using a phrase that's more likely to appear in documents discussing both concepts. Maybe "Latino and white identity intersection" or "Can someone be both white and Latino." \textcolor{gray}{\{... More verbose reasoning omitted\}...}

\textcolor{purple}{After considering these options, the best approach is to \textbf{({\large \texttwemoji{trophy}}) rephrase  the question into a statement} that clearly states the intersection, which is likely to be covered in relevant documents. So, the rewritten query might be "Can a person be both white and Latino?" which is similar to the original \textbf{but phrased as a statement that Contriever can retrieve documents about.}}
\textcolor{purple}{</think>}

\textcolor{blue}{\textbf{<rewrite> Being white and Latino </rewrite>}}

-----------------

({\large \texttwemoji{rocket}}) \textcolor{teal}{\textbf{Result Metric: nDCG@10 of 0.965}} 

\end{varwidth}
    }

    \caption{Novel Strategy Discovery by rewriting agents. Emojis are annotated with meanings:  {\large \texttwemoji{thinking face}}: Retriever-aware reflections. {\large \texttwemoji{light bulb}}: Policy Proposal. {\large \texttwemoji{trophy}}: Policy Confirmation. {\large \texttwemoji{rocket}}: Successful Retrieval which yields advantage in GRPO rollouts.
    }
    \label{fig:contriever_case_study}
    \vspace{-15pt}
\end{figure}

\begin{table*}[ht]
    \centering
    \resizebox{0.96\textwidth}{!}{%
        \begin{tabular}{lccccccccccccc}
            \toprule
            \textbf{Retriever} & \textbf{CovidQA} & \textbf{DelucionQA} & \textbf{EManual} & \textbf{ExpertQA} & \textbf{FinQA} & \textbf{HAGRID} & \textbf{HotpotQA} & \textbf{MS Marco} & \textbf{PubmedQA} & \textbf{TAT-QA} & \textbf{TechQA} & \textbf{Avg.} \\
            \midrule
            Contriever  & 36.0 & 61.6 & 55.1 & 52.5 & 14.4 & 68.9 & 43.8 & 72.0 & 62.3 & 8.6 & 24.7 &  45.5\\
            Contriever + rewrite general         & 40.4 & 70.0 & 65.7 & 60.8 & 21.0 & 85.8 & 64.1 & 77.2 & 71.4 & 13.6 & 30.0 & 54.5$\uparrow$\\
            Contriever + rewrite direct          & 42.1 & 71.3 & 71.1 & 62.6 & 23.3 & 87.2 & 63.9 & 78.1 & 72.9 & 14.9 & 31.9 & 56.3$\uparrow$\\
            Contriever + rewrite explore         & 41.9 & \textbf{72.1} & 65.2 & 54.7 & 18.1 & 82.4 & 63.9 & 77.2 & 68.5 & 12.3 & 27.9 &  53.2$\uparrow$\\
            Contriever + rewrite general + turn2 & 41.8 & 68.7 & 62.8 & \textbf{64.0} & 22.9 & \textbf{87.9} & \textbf{68.1} & 76.0 & 69.9 & 13.9 & 30.9 & 55.2$\uparrow$\\
            Contriever + rewrite direct + turn2  & \textbf{42.6} & 70.1 & \textbf{71.5} & 63.5 & \textbf{23.3} & 86.1 & 64.9 & \textbf{78.5} & \textbf{73.0} & \textbf{15.0} & \textbf{32.5} & \textbf{56.5$\uparrow$} \\
            \midrule
            BM25   & 30.8 & 57.3 & 40.0 & 50.6 & 20.0 & 67.3 & 68.7 & 57.9 & 48.1 & 10.4 & 44.0 &  45.0\\
            BM25 + rewrite general       & 37.6 & 66.1 & 56.1 & 49.2 & 29.3 & 74.9 & 71.1 & 69.8 & 58.6 & 21.5 & 49.7 & 53.1$\uparrow$\\
            BM25 + rewrite explore          & \textbf{40.5} & \textbf{66.1} & \textbf{66.3} & \textbf{46.1} & \textbf{33.4} & \textbf{77.9} & \textbf{82.6} & \textbf{72.5} & \textbf{61.9} & \textbf{24.9} & \textbf{53.7} & \textbf{56.9$\uparrow$} \\
            BM25 + rewrite direct        & 40.2 & 66.0 & 66.2 & 47.0 & 31.3 & 77.4 & 82.2 & 71.3 & 61.7 & 24.4 & 50.1 & 56.2 \\
            \midrule
            all-MiniLM-L6-v2         & 39.7 & 68.1 & 65.5 & 61.2 & 11.1 & 84.4 & 58.8 & 79.3 & 72.2 & 11.2 & 30.3 & 52.9\\
            all-MiniLM-L6-v2 + rewrite general & \textbf{41.4} & \textbf{72.1} & \textbf{71.1} & \textbf{61.2} & \textbf{13.5} & \textbf{86.4} & \textbf{61.1} & \textbf{81.8} & \textbf{75.4} & \textbf{13.7} & \textbf{31.4} & \textbf{55.4$\uparrow$}\\
            \midrule
            Qwen3-Embedding-0.6B           & 56.3 & 77.8 & 73.7 & 66.7 & 27.1 & 89.2 & 67.0 & 81.0 & 79.1 & 17.6 & 41.2 & 61.5\\
            Qwen3-Embedding-0.6B + rewrite general & \textbf{60.5} & \textbf{83.4} & \textbf{77.4} & \textbf{66.9} & \textbf{30.0} & \textbf{89.4} & \textbf{68.0} & \textbf{84.5} & \textbf{83.2} & \textbf{20.1} & \textbf{41.2} & \textbf{64.1$\uparrow$}\\
            \bottomrule
        \end{tabular}%
    }
    \caption{Results of LLM query rewriters with different retrievers on the RAGBench.}
    \label{RagBench}
\end{table*}

\begin{table*}[ht]
    \centering
    \resizebox{\textwidth}{!}{%
        \begin{tabular}{lccccccccccccccc}
            \toprule
            \textbf{Retriever} & 
            \textbf{ArguAna} &
            \textbf{MSMARCO} & \textbf{TREC-COVID} & \textbf{NFCorpus} & \textbf{NQ} & \textbf{HotpotQA} & \textbf{FiQA} &  \textbf{Touché} & \textbf{Quora} & \textbf{DBPedia} & \textbf{SCIDOCS} & \textbf{FEVER} & \textbf{Climate-FEVER} & \textbf{SciFact} & \textbf{Avg.} \\
            \midrule
            Contriever   & 33.55 & 36.62 & 17.36 & 27.11 & 18.05 & 41.01 & 12.41 & 6.75 & 83.36 & 25.08 & 10.97 & 27.22 & 7.16 & 57.14 & 28.84 \\ 
             \textbf{w/ Our rewriter} & 32.75 & \textbf{\textcolor{ForestGreen}{49.23$\uparrow$}} & \textbf{\textcolor{ForestGreen}{32.08$\uparrow$}} & 26.71 & \textbf{\textcolor{ForestGreen}{36.28$\uparrow$}} & \textbf{\textcolor{ForestGreen}{52.44$\uparrow$}} & \textbf{\textcolor{ForestGreen}{16.17$\uparrow$}} & \textbf{\textcolor{ForestGreen}{7.68$\uparrow$}} & 77.27 & \textbf{\textcolor{ForestGreen}{32.18$\uparrow$}} & \textbf{\textcolor{ForestGreen}{13.10$\uparrow$}} & \textbf{\textcolor{ForestGreen}{37.52$\uparrow$}} & \textbf{\textcolor{ForestGreen}{11.75$\uparrow$}} & \textbf{\textcolor{ForestGreen}{64.54$\uparrow$}} & \textbf{\textcolor{ForestGreen}{34.98$\uparrow$}} \\
            \midrule
            w/ Gemini-2.5-flash \\
            - general prompt & 31.21 & 38.34 & 28.21 & 26.58 & 16.36 & 39.16 & 10.33 & 4.58 & 68.99 & 23.38 & 10.55 & 22.34 & 6.21 & 51.31 & 26.97 \\
            - specific Contriever prompt  & 30.72 & 37.08 & 28.13 & 26.25 & 18.70 & 39.40 & 10.13 & 3.85 & 64.87 & 25.70 & 10.79 & 24.77 & 6.47 & 56.70 & 27.40 \\
        \bottomrule
        \end{tabular}%
    }
    \caption{Zero-shot Adaptation of LLM query rewriters to the BEIR benchmark.}
    \label{tab:beir_results}
\end{table*}

\begin{table}[ht]
    \centering
    \resizebox{0.9\linewidth}{!}{\begin{tabular}{lc}
        \toprule
        \textbf{Retriever} & \textbf{FinAgentBench} \\
        \midrule
        Contriever                           & 6.43 \\
        Contriever + rewrite                 & \hspace{4.2pt}\textbf{\textcolor{ForestGreen}{7.39$\uparrow$}} \\
        \midrule
        BM25                                 & 8.17 \\
        BM25 + rewrite                       & \hspace{4.2pt}\textbf{\textcolor{ForestGreen}{9.02$\uparrow$}}  \\
        \midrule
        all-MiniLM-L6-v2                     & 6.69 \\
        all-MiniLM-L6-v2 + rewrite           & \hspace{4.2pt}\textbf{\textcolor{ForestGreen}{7.17$\uparrow$}}\\
        \midrule
        Qwen3-Embedding-0.6B                 & 9.10 \\
        Qwen3-Embedding-0.6B + rewrite       & \hspace{4.5pt}\textbf{\textcolor{ForestGreen}{10.81$\uparrow$}} \\
        \bottomrule
    \end{tabular}}
    \caption{nDCG@10 performance on FinAgentBench.}
    \label{tab:finagent_results}
    \vspace{-5pt}
\end{table}

\paragraph{Infrastructure} We serve the retrieval environments (retrievers and pre-indexed corpus) as APIs using \texttt{FastAPI} and \texttt{uvicorn}. The retrieval environments are served on different nodes, isolated from the main GRPO training process, which secures no interference with the RL training.

\paragraph{RL implementation}

For multi-turn search, we assign 0.5 weighting to the first-round retrieval and 1 to the second-sround retrieval. 
We apply masking to the search results returned by the retrievers, and only compute the loss of the reasoning and the rewritten queries. This is based on the principle of not punishing or rewarding a model for texts it didn't generate. We further ablate whether it is more optimal to apply $n^2$ rollouts to the full unconditional two-turn search trajectories, or $n$ rollouts to the second-turn rewriting conditioned on $n$ first-turn rewriting and search results.

\paragraph{Retrievers} We select four representative retrieval systems, including \texttt{Qwen3-Embedding}~\cite{zhang2025qwen3}, \texttt{all-MiniLM-L6-v2}~\cite{sentence-transformers_all-MiniLM-L6-v2_2025}, \texttt{Contriever}~\cite{contriever}, and \texttt{BM25}~\cite{DBLP:journals/ftir/BM25RobertsonZ09}.

\texttt{Qwen3-Embedding} and \texttt{all-MiniLM-L6-v2} are dense embedding models trained on large-scale supervised pairs, and they respectively represent decoder/high-capability/large and encoder/low-capability/small models. 

\texttt{Contriever} is an unsupervised representation model trained on randomly cropped spans of Wikipedia article as positive pairs. Therefore, Contriever intuitively prefers retrieving documents using document-style query, while providing surpar performance using question-style query.

\texttt{BM25} is a sparse retrieval algorithm, which in its nature prefers keyword-style retrieval. When the metadata (e.g., inverse document frequency) is computed on non-massive corpus, it is also sensitive to small-variants of phrasing and spelling (e.g., carryforward vs. carryforwards).

We follow the optimal encoding setting of all retrieval systmes, such as using mean pooling for Contriever, and using the query instruction template for Qwen3-Embedding, etc.

\section{Results}
\label{sec:results}

\subsection{Main Results}
\paragraph{RQ1.1: Can LLMs adapt their query formulation strategies to local retrievers via RL?}
We observe significant performance gains across diverse retrievers on the in-distribution RAGBench dataset (Table~\ref{RagBench}). Adapting query formulation to the specific retriever—whether through general prompts, specific instructions, or exploratory RL—consistently outperforms the baseline usage of the retriever alone, e.g., the BM25 retriever sees a substantial jump in nDCG@10 when using the ``explore'' rewriting strategy compared to its raw baseline. This confirms that the model successfully learns distinct ``dialects'' suited for different backends.

The learned query rewriter LLM agents also demonstrate strong generalization capabilities when applied to out-of-distribution and domain-specific environments. On the BEIR benchmark (Table~\ref{tab:beir_results}), our query rewriter improves the average performance of Contriever from 28.84 to 34.98 in a zero-shot setting.
In contrast, Gemini-2.5-Flash fails to provide useful rewritten queries even if under the specific prompts aligned with our rewriters, resulting in performance drop.

Furthermore, results on the specialized FinAgentBench (Table~\ref{tab:finagent_results}) show universal improvements across all tested systems—including Contriever, BM25, and Qwen3-Embedding—proving that the learned structural adaptations remain effective even under significant domain shifts.

Although our primary focus is on isolating the retrieval component from the full RAG pipeline, improved retrieval quality directly increases the likelihood that the gold document ($d^*$) is ranked highly, thereby improving downstream answer accuracy. In Appendix~\ref{apx:RAG_performance}, we empirically verify that these retrieval gains translate into improvements in end-to-end RAG performance.

\subsection{Measuring the Discrepancy of Optimal Queries}
\paragraph{RQ1.2: Can strategies learned for one retriever transferable to others?}
\label{sec:remmd}

\begin{figure}[ht]
    \centering
    \includegraphics[width=0.8\linewidth]{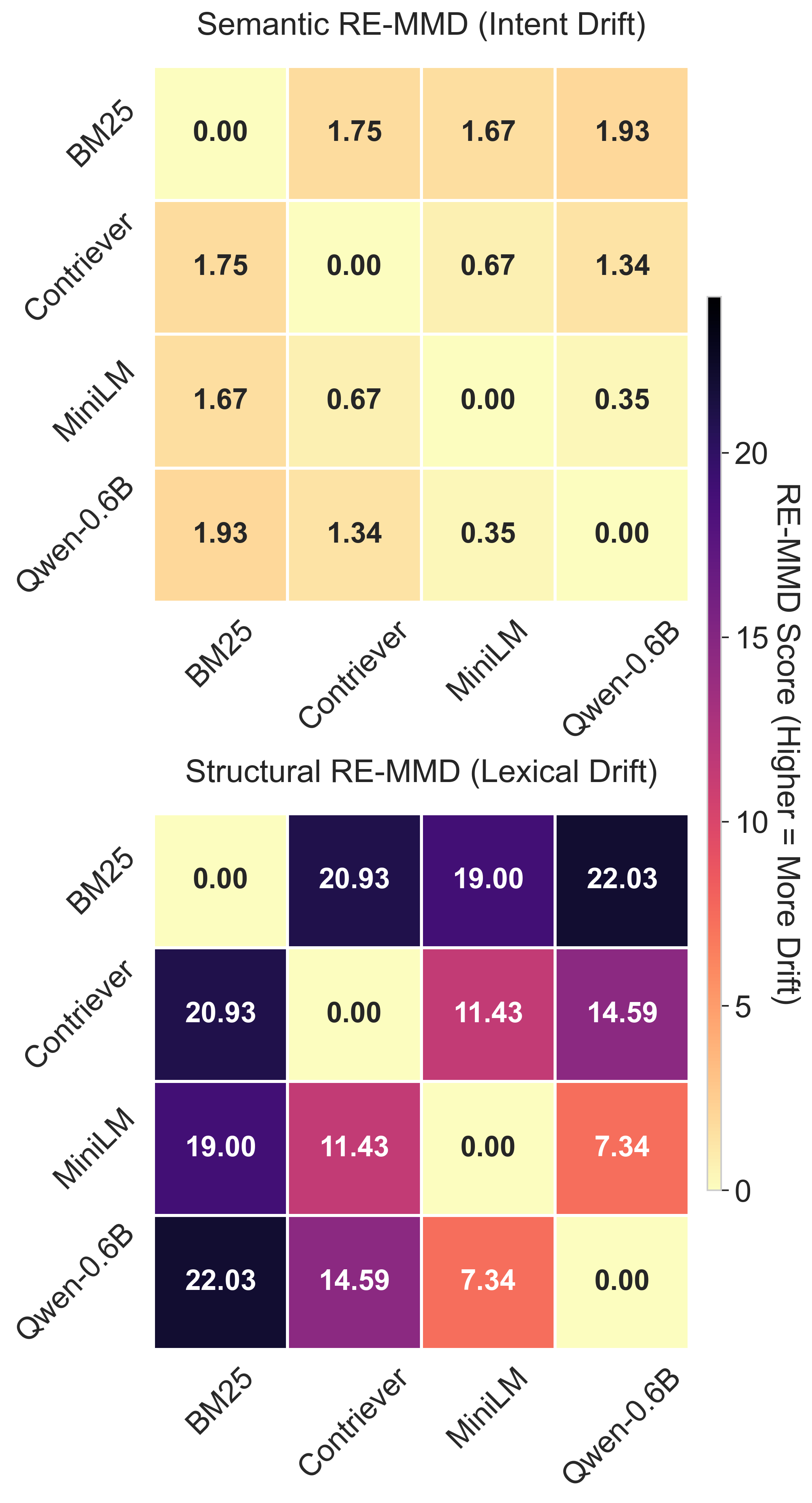}
    \caption{Measuring semantic and structural drifts of optimal queries across retrieval environments. Scores are our proposed RE-MMD values multiplied by 100.}
    \label{fig:re-mmd}
\end{figure}

To quantify the transferability of learned query formulation strategies across retrievers, we introduce \textbf{Retrieval Environment MMD (RE-MMD)}, a metric based on Maximum Mean Discrepancy that measures the distance between the distributions of optimal queries in a source retrieval environment $\mathcal{E}_{\text{src}}$ and a target environment $\mathcal{E}_{\text{tgt}}$. We decompose distributional discrepancies into two complementary types of drift:
\begin{itemize}[leftmargin=*,topsep=2pt]
\setlength{\itemsep}{2pt}
\setlength{\parskip}{0pt}
\setlength{\parsep}{0pt}
\item \textbf{Semantic Drift (Intent).} Do the queries target different underlying information needs? We measure this using a dense semantic encoder $\phi_{\text{sem}}$ (e.g., MiniLM) that maps queries into an intent representation space.
\item \textbf{Structural Drift (Formulation).} Do the queries differ in their surface-level formulation or “dialect” (e.g., keyword-based versus natural-language expressions)? We capture this using lexical representations $\phi_{\text{struct}}$.
\end{itemize}

\begin{figure}[ht]
    \includegraphics[width=0.92\linewidth]{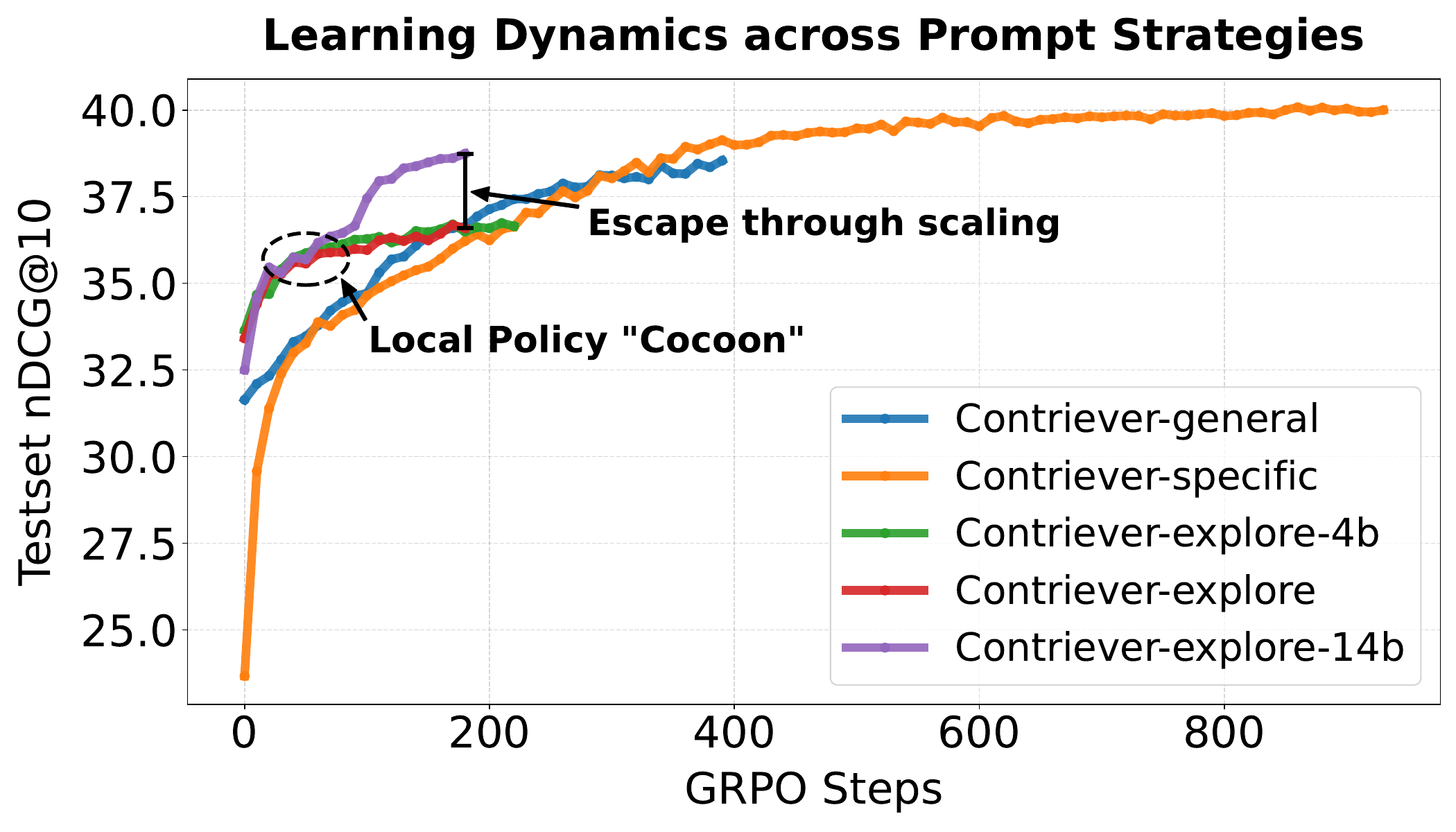}
    \vspace{0.5cm}    \includegraphics[width=0.92\linewidth]{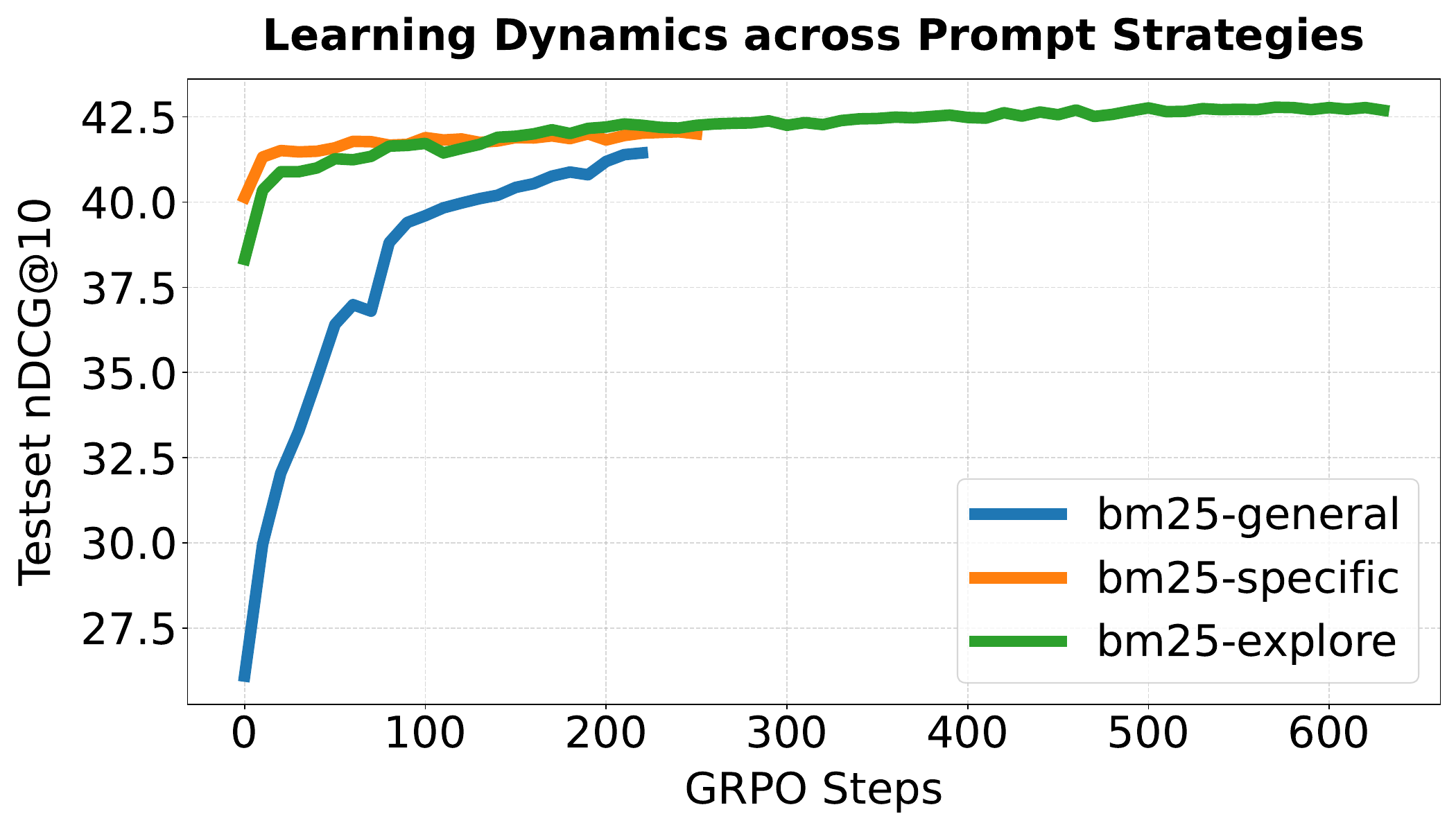}
    \caption{Learning dynamics across different prompt methods and model sizes of query rewriter LLM. 
    }    \label{fig:learning_dynamics_across_prompts}
\vspace{-0.15in}
\end{figure}

Formally, we define the squared RE-MMD distance in a Reproducing Kernel Hilbert Space (RKHS) $\mathcal{H}$ as: $\text{RE-MMD}^2(\mathcal{E}_{\text{src}}, \mathcal{E}_{\text{tgt}}) = \left\| \mu_{\text{src}} - \mu_{\text{tgt}} \right\|_{\mathcal{H}}^2$,
where $\mu = \mathbb{E}_{q \sim \mathcal{P}}[\phi(q)]$ denotes the expected embedding of successful queries in corresponding environments. Details of the empirical estimation using kernel-based methods are in Appendix~\ref{app:remmd_details}.

\paragraph{Results.}
Figure~\ref{fig:re-mmd} reports the RE-MMD scores (scaled by 100) across retriever pairs. We observe a clear contrast between the two types of drift. \textbf{Semantic RE-MMD remains consistently low} (typically $<2.0$), indicating that the underlying information need is largely preserved across retrievers. In contrast, \textbf{Structural RE-MMD is an order of magnitude larger} (often $>20.0$), especially between sparse retrievers (e.g., BM25) and dense retrievers (e.g., Contriever). These results provide quantitative evidence that strategy transfer failures are primarily \emph{stylistic} rather than \emph{intentional}: the agent fails not because it seeks different information, but because it cannot effectively “code-switch” into the query formulation style—such as keyword density versus declarative phrasing—required by the target retrieval environment. Overall, \textit{our findings suggest that query formulation policies optimized for one retriever generally do not transfer reliably to others}.

\subsection{Human Guidance in Prompt}
\paragraph{RQ2: How does the comprehensiveness of human guidance in prompts affect the learning of query formulation?}

Prompt design is known to substantially influence the learning dynamics of GRPO-based optimization. In this section, we investigate how different levels of human guidance embedded in prompts shape the learning behavior of query formulation policies. Specifically, we analyze learning dynamics under different prompt priors that vary in the amount of retriever-specific knowledge provided to the model.

We conduct in-depth experiments in retrieval environments built with \texttt{Contriever} and \texttt{bm25}, which represent two retriever classes with distinct and human-interpretable optimal query behaviors. Owing to its {document span, document span} contrastive training objective, \texttt{Contriever} is expected to perform best when the original query is transformed into a hypothetical document resembling the target document. In contrast, the statistical nature of \texttt{bm25} inherently favors concise, keyword-oriented queries.

We design three levels of prompt comprehensively (Figure~\ref{fig:training_prompts} in Appendix~\ref{apx: training prompts} provides the prompts in detail):
\begin{itemize}[leftmargin=*,topsep=2pt]
\setlength{\itemsep}{2pt}
\setlength{\parskip}{0pt}
\setlength{\parsep}{0pt}
    \item \textbf{General Prompts.} We provide general query rewriting guidelines along with five candidate rewriting strategies. The query rewriter is instructed to reason over these options and select one strategy per rollout. In this setting, the rewriter is not informed of the underlying retriever and must infer its behavior solely through GRPO reward signals during training.
    \item \textbf{Exploratory Prompts.} We inform the query rewriter of the retriever being used and provide a high-level description of how it was trained. The rewriter is encouraged to explore and infer an optimal query formulation strategy based on its understanding of the retriever’s behavior.
    \item \textbf{Specific Prompts.} We explicitly inform the query rewriter of the retriever and prescribe a concrete query formulation strategy provided from human experts, e.g., hypothetical document generation for \texttt{Contriever} and keyword-based rewriting for \texttt{bm25}. The rewriter is instructed to apply only this retriever-specific strategy.
\end{itemize}

Figure~\ref{fig:learning_dynamics_across_prompts} illustrates the learning dynamics for \texttt{Contriever}. The {general prompt} achieves a reasonable initial performance but improves slowly, as the large space of rewriting choices makes exploration inefficient. The {exploratory prompt} yields the strongest initial performance and rapid improvement within the first 50 training steps. However, we observe that the model quickly converges to a suboptimal policy that emphasizes keyword matching, which it prematurely identifies as effective for \texttt{Contriever}. This early policy collapse leads to convergence around the 200th step. In contrast, the {specific prompt}, which enforces hypothetical document generation, achieves the best final performance. Although this setting starts from a relatively weaker initial point—due to the rewriter’s random guesses about the structure of target documents in the black-box corpus—it benefits from consistent guidance and gradually refines its strategy through GRPO rollouts, ultimately exhibiting higher performance and slower convergence.

The conclusions differ for \texttt{bm25}, where the {exploratory prompt} outperforms the {specific prompt}, with both substantially outperforming the {general prompt}. Notably, even under exploratory prompts, the rewriter infers early in training that keyword-style queries are optimal for \texttt{bm25}. This can be attributed to \texttt{bm25} being a well-established retrieval algorithm whose underlying principles are extensively discussed in the pretraining corpora of LLMs. In contrast, general prompts yield the poorest initial performance, as \texttt{bm25}'s sensitivity to lexical choices causes many suboptimal rewrites to largely degrade early-stage rewards.

Taken together, these seemingly contrasting results reveal an intuitive principle: \textit{when an LLM lacks sufficient prior knowledge or reasoning capability about a retriever’s optimal query strategy, explicit human expert guidance can effectively steer RL exploration and improve learning outcomes. Conversely, when the retriever’s behavior is already well captured in the model’s pretraining knowledge, exploratory prompts are sufficient—and can even outperform rigid, human-prescribed strategies.}

\begin{figure*}[t] 
    \centering
    \includegraphics[width=0.8\linewidth]{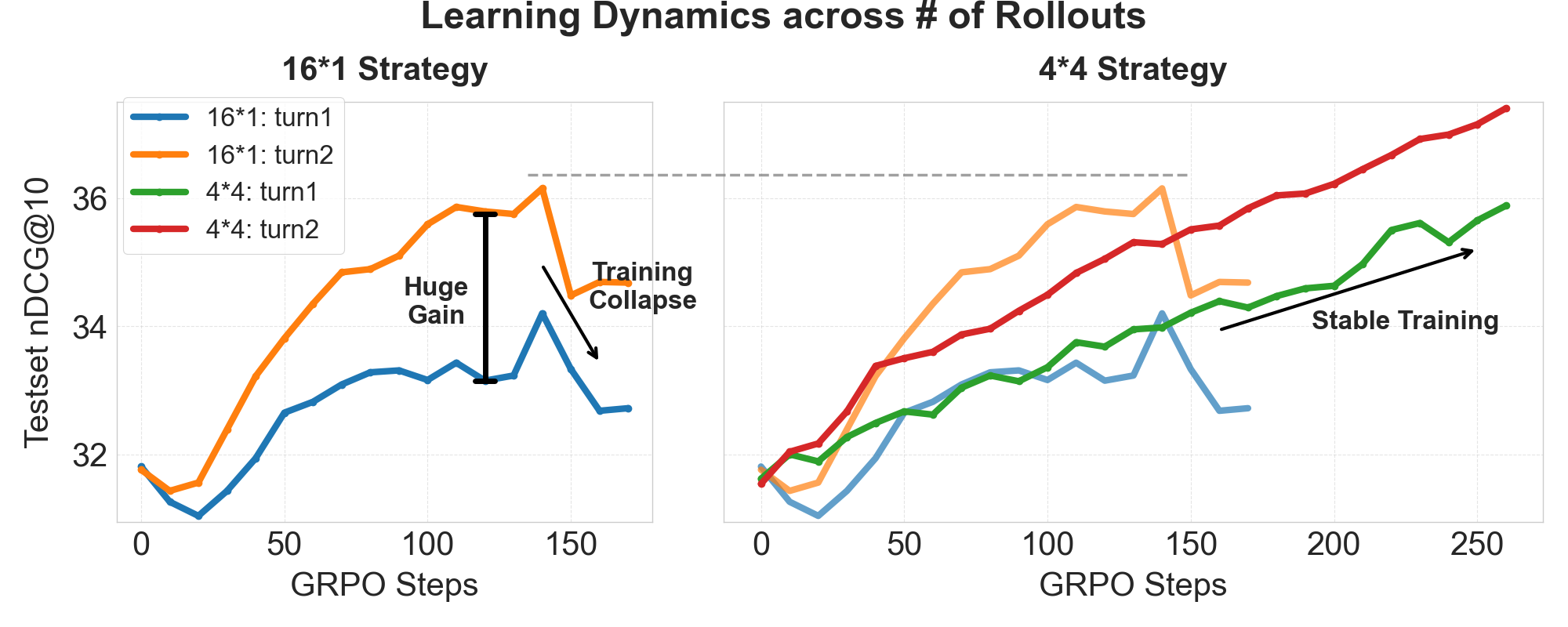}
    \caption{Training dynamics comparison of independent and branching rollout techniques}
    \label{fig:multi}
\end{figure*}

\subsection{Scaling of Model Size and Retrieval Steps}
\paragraph{RQ3.1: Can we enhance query formulation performance by scaling up model size?}

We examine the effect of LLM model size on query formulation performance. As shown in Figure~\ref{fig:learning_dynamics_across_prompts}, increasing the rewriter size to 14B enables the model to break through the performance plateaus observed for the 4B and 8B variants. This behavior highlights the role of increased model capacity and world knowledge in enabling more diverse and effective exploration, thereby mitigating premature policy collapse during RL training.

Concretely, while prior knowledge—either learned implicitly (e.g., the 8B model discovering that longer queries are beneficial under general prompts) or provided explicitly via human expert guidance (e.g., document-style queries for \texttt{Contriever})—suggests that longer or document-like queries are optimal, the 14B model goes beyond these assumptions. Under exploratory prompts, it discovers a distinct strategy that can be characterized as \emph{statement-style} queries, which differ from both question-form and document-style rewrites. For example, given the original query \textit{Is it possible to be white and Latino?''}, a question-style rewrite such as \textit{Can someone be both white and Latino?''} achieves an nDCG@10 of 0.542, whereas a concise statement-style rewrite—\textit{``Being white and Latino''}—surprisingly attains an nDCG@10 of 0.965. This strategy emerges naturally from RL training with a 14B model under exploratory prompts, as illustrated in Figure~\ref{fig:contriever_case_study}.

We further observe that the diversity of strategy exploration scales with model size. At the early stages of training, larger models exhibit lower initial average rewards, reflecting broader exploration over a wider range of candidate rewriting strategies, many of which are initially suboptimal. This increased exploratory behavior, however, ultimately enables the discovery of more effective query formulation policies.

Overall, the model scaling results reinforce insights from the prompt ablation studies in the previous section. When base models lack sufficient internal knowledge about retriever-specific behaviors, human expert–driven prompts can effectively constrain the search space and guide learning toward strong policies. In contrast, \textit{larger and more capable models are able to discover non-obvious strategies that even human experts may not anticipate, allowing them to escape the local policy optima that constrain smaller models such as the 4B and 8B variants.}

\paragraph{RQ3.2: Can we enhance query formulation performance by scaling up retrieval steps?}
We observe that the {${4\times4}$ Strategy} (branching) significantly outperforms the {${16\times 1}$ Strategy} (independent) in training stability. This performance gap stems from how the branching structure solves the credit assignment problem inherent in our shared reward function ($R_{\text{total}} = 0.5 R_{\text{turn1}} + R_{\text{turn2}}$).

The branching strategy isolates the learning signal for the second turn. By generating $k=4$ continuations from a fixed first-round outcome, the first-round reward becomes a constant $C$ within that subgroup. When calculating the advantage for the second turn, this constant cancels out:
\begin{equation}
\small
    A_{\text{turn2}}^{(i)} \approx (0.5 C + R_{\text{turn2}}^{(i)}) - \mathbb{E}_{\text{siblings}}[0.5 C + R_{\text{turn2}}]\text{.}
\end{equation}
This effectively removes the noise from the first turn, allowing the model to receive a precise gradient based solely on the quality of the second turn.

Simultaneously, the $4\times4$ strategy stabilizes the first turn by acting as a Monte Carlo estimator for its value. In the $16\times1$ strategy, the "future value" of a first-round rewrite is based on a single, noisy second-round sample. 
In contrast, the $4\times4$ strategy estimates the value of a first-round action by adding its immediate reward to the average reward of the subsequent four second-round attempts that follow it.

This tells the model: \textit{"Given this specific first-round attempt, here is the expected outcome if we try 4 different second-round attempts."} This averaging reduces variance in the reward signal for the first turn, preventing the training collapse observed in the independent baseline.

\section{Related Work}

Retrieval-augmented generation (RAG) is a standard approach for augmenting LLMs on tasks that exceed LLMs' knowledge boundaries \cite{lewis2020retrieval,xiao2025analyzing}. For bridging the behavioral gaps between LLMs and retrievers, prior approaches mitigate semantic mismatches via query rewriting \cite{ma2023query,chan2024rq} or zero-shot expansion \cite{gao2023precise}. While Reinforcement Learning has been used to align rewrites with generation goals \cite{nogueira2017task,ma2023query,jin2025search}, these methods typically assume a uniform retrieval environment. On the information retrieval side, prior works show that different retrievers display distinct length preferences \cite{thakur2021beir,xiao2023length}. In this work, we systematically analyze \textit{retriever-specific} query adaptation, addressing the ``structural drift'' where policies tailored for one backend fail to transfer to others.

Agentic frameworks like Self-RAG \cite{asai2024selfrag} and Adaptive-RAG \cite{mallen2023not} utilize dynamic, multi-turn retrieval but often face training instability in long-horizon trajectories \cite{shao2023enhancing}. Complementing research on the retriever-generator preference gap \cite{ke2024bridging}, we introduce a branching rollout technique that effectively stabilizes the optimization of multi-step query formulation strategies.

\section{Conclusion}
We present the first systematic study on adapting LLM query formulation to specific retrievers via Reinforcement Learning. We demonstrate that optimal strategies are highly retriever-dependent and quantify this stylistic transfer failure as ``Structural Drift.'' Furthermore, we find that scaling model size enables the discovery of non-intuitive strategies that outperform expert heuristics. Finally, our proposed branching rollout technique stabilizes multi-turn training, paving the way for robust, truly retriever-aware RAG systems.
\section*{Limitations}
First, our study focuses exclusively on text-only retrievers and text-based document collections, and does not consider multimodal retrievers \cite{xiao2025scaling,chia2025m,xiao2025mieb} that can retrieve heterogeneous content such as images or audio. Extending our analysis to multimodal retrieval settings is an important direction for future work. Second, in our experiments on scaling the number of retrieval steps, we limit our evaluation to one-step and two-step retrieval. While these settings are sufficient to demonstrate the benefits of multi-step retrieval, it would be valuable to investigate learning dynamics and performance trends under longer retrieval trajectories in future studies.

\bibliography{custom}

@article{friel2024ragbench,
  title={Ragbench: Explainable benchmark for retrieval-augmented generation systems},
  author={Friel, Robert and Belyi, Masha and Sanyal, Atindriyo},
  journal={arXiv preprint arXiv:2407.11005},
  year={2024}
}

@inproceedings{choi2025finagentbench,
  title={Finagentbench: A benchmark dataset for agentic retrieval in financial question answering},
  author={Choi, Chanyeol and Kwon, Jihoon and Lopez-Lira, Alejandro and Kim, Chaewoon and Kim, Minjae and Hwang, Juneha and Ha, Jaeseon and Choi, Hojun and Yun, Suyeol and Kim, Yongjin and others},
  booktitle={Proceedings of the 6th ACM International Conference on AI in Finance},
  pages={632--637},
  year={2025}
}

@article{DBLP:journals/ftir/BM25RobertsonZ09,
  author       = {Stephen E. Robertson and
                  Hugo Zaragoza},
  title        = {The Probabilistic Relevance Framework: {BM25} and Beyond},
  journal      = {Found. Trends Inf. Retr.},
  volume       = {3},
  number       = {4},
  pages        = {333--389},
  year         = {2009},
  url          = {https://doi.org/10.1561/1500000019},
  doi          = {10.1561/1500000019},
  bibsource    = {dblp computer science bibliography, https://dblp.org}
}

@inproceedings{moller2020covid,
  title={COVID-QA: A question answering dataset for COVID-19},
  author={M{\"o}ller, Timo and Reina, Anthony and Jayakumar, Raghavan and Pietsch, Malte},
  booktitle={Proceedings of the 1st Workshop on NLP for COVID-19 at ACL 2020},
  year={2020}
}

@article{contriever,
  author       = {Gautier Izacard and
                  Mathilde Caron and
                  Lucas Hosseini and
                  Sebastian Riedel and
                  Piotr Bojanowski and
                  Armand Joulin and
                  Edouard Grave},
  title        = {Unsupervised Dense Information Retrieval with Contrastive Learning},
  journal      = {Trans. Mach. Learn. Res.},
  volume       = {2022},
  year         = {2022},
  url          = {https://openreview.net/forum?id=jKN1pXi7b0},
  bibsource    = {dblp computer science bibliography, https://dblp.org}
}

@inproceedings{jin2019pubmedqa,
  title={Pubmedqa: A dataset for biomedical research question answering},
  author={Jin, Qiao and Dhingra, Bhuwan and Liu, Zhengping and Cohen, William and Lu, Xinghua},
  booktitle={Proceedings of the 2019 conference on empirical methods in natural language processing and the 9th international joint conference on natural language processing (EMNLP-IJCNLP)},
  pages={2567--2577},
  year={2019}
}

@inproceedings{yang2018hotpotqa,
  title={HotpotQA: A dataset for diverse, explainable multi-hop question answering},
  author={Yang, Zhilin and Qi, Peng and Zhang, Saizheng and Bengio, Yoshua and Cohen, William and Salakhutdinov, Ruslan and Manning, Christopher D},
  booktitle={Proceedings of the 2018 conference on empirical methods in natural language processing},
  pages={2369--2380},
  year={2018}
}

@article{bajaj2016ms,
  title={Ms marco: A human generated machine reading comprehension dataset},
  author={Bajaj, Payal and Campos, Daniel and Craswell, Nick and Deng, Li and Gao, Jianfeng and Liu, Xiaodong and Majumder, Rangan and McNamara, Andrew and Mitra, Bhaskar and Nguyen, Tri and others},
  journal={arXiv preprint arXiv:1611.09268},
  year={2016}
}

@inproceedings{
hendrycks2021cuad,
title={{CUAD}: An Expert-Annotated {NLP} Dataset for Legal Contract Review},
author={Dan Hendrycks and Collin Burns and Anya Chen and Spencer Ball},
booktitle={Thirty-fifth Conference on Neural Information Processing Systems Datasets and Benchmarks Track (Round 1)},
year={2021},
url={https://openreview.net/forum?id=7l1Ygs3Bamw}
}

@inproceedings{nandy-etal-2021-question-answering,
    title = "Question Answering over Electronic Devices: A New Benchmark Dataset and a Multi-Task Learning based {QA} Framework",
    author = "Nandy, Abhilash  and
      Sharma, Soumya  and
      Maddhashiya, Shubham  and
      Sachdeva, Kapil  and
      Goyal, Pawan  and
      Ganguly, NIloy",
    editor = "Moens, Marie-Francine  and
      Huang, Xuanjing  and
      Specia, Lucia  and
      Yih, Scott Wen-tau",
    booktitle = "Findings of the Association for Computational Linguistics: EMNLP 2021",
    month = nov,
    year = "2021",
    address = "Punta Cana, Dominican Republic",
    publisher = "Association for Computational Linguistics",
    url = "https://aclanthology.org/2021.findings-emnlp.392/",
    doi = "10.18653/v1/2021.findings-emnlp.392",
    pages = "4600--4609",
}

@inproceedings{castelli2020techqa,
  title={The techqa dataset},
  author={Castelli, Vittorio and Chakravarti, Rishav and Dana, Saswati and Ferritto, Anthony and Florian, Radu and Franz, Martin and Garg, Dinesh and Khandelwal, Dinesh and McCarley, J Scott and McCawley, Michael and others},
  booktitle={Proceedings of the 58th annual meeting of the association for computational linguistics},
  pages={1269--1278},
  year={2020}
}

@inproceedings{chen2021finqa,
  title={Finqa: A dataset of numerical reasoning over financial data},
  author={Chen, Zhiyu and Chen, Wenhu and Smiley, Charese and Shah, Sameena and Borova, Iana and Langdon, Dylan and Moussa, Reema and Beane, Matt and Huang, Ting-Hao and Routledge, Bryan R and others},
  booktitle={Proceedings of the 2021 Conference on Empirical Methods in Natural Language Processing},
  pages={3697--3711},
  year={2021}
}

@inproceedings{zhu2021tat,
  title={TAT-QA: A Question Answering Benchmark on a Hybrid of Tabular and Textual Content in Finance},
  author={Zhu, Fengbin and Lei, Wenqiang and Huang, Youcheng and Wang, Chao and Zhang, Shuo and Lv, Jiancheng and Feng, Fuli and Chua, Tat-Seng},
  booktitle={Proceedings of the 59th Annual Meeting of the Association for Computational Linguistics and the 11th International Joint Conference on Natural Language Processing (Volume 1: Long Papers)},
  pages={3277--3287},
  year={2021}
}

@inproceedings{malaviya2024expertqa,
  title={Expertqa: Expert-curated questions and attributed answers},
  author={Malaviya, Chaitanya and Lee, Subin and Chen, Sihao and Sieber, Elizabeth and Yatskar, Mark and Roth, Dan},
  booktitle={Proceedings of the 2024 Conference of the North American Chapter of the Association for Computational Linguistics: Human Language Technologies (Volume 1: Long Papers)},
  pages={3025--3045},
  year={2024}
}

@article{kamalloo2023hagrid,
  title={Hagrid: A human-llm collaborative dataset for generative information-seeking with attribution},
  author={Kamalloo, Ehsan and Jafari, Aref and Zhang, Xinyu and Thakur, Nandan and Lin, Jimmy},
  journal={arXiv preprint arXiv:2307.16883},
  year={2023}
}

@misc{sentence-transformers_all-MiniLM-L6-v2_2025,
  title        = {{all-MiniLM-L6-v2}},
  author       = {{Sentence-Transformers}},
  year         = {2021},
  howpublished = {\url{https://huggingface.co/sentence-transformers/all-MiniLM-L6-v2}},
  note         = {Accessed: 2026-01-06}, 
  organization = {Hugging Face},
}

@article{zhang2025qwen3,
  title={Qwen3 Embedding: Advancing Text Embedding and Reranking Through Foundation Models},
  author={Zhang, Yanzhao and Li, Mingxin and Long, Dingkun and Zhang, Xin and Lin, Huan and Yang, Baosong and Xie, Pengjun and Yang, An and Liu, Dayiheng and Lin, Junyang and others},
  journal={arXiv preprint arXiv:2506.05176},
  year={2025}
}

@inproceedings{ma2023query,
  title={Query rewriting in retrieval-augmented large language models},
  author={Ma, Xinbei and Gong, Yeyun and He, Pengcheng and Zhao, Hai and Duan, Nan},
  booktitle={Proceedings of the 2023 Conference on Empirical Methods in Natural Language Processing},
  pages={5303--5315},
  year={2023}
}

@inproceedings{gao2023precise,
  title={Precise zero-shot dense retrieval without relevance labels},
  author={Gao, Luyu and Ma, Xueguang and Lin, Jimmy and Callan, Jamie},
  booktitle={Proceedings of the 61st Annual Meeting of the Association for Computational Linguistics (Volume 1: Long Papers)},
  pages={1762--1777},
  year={2023}
}

@article{chan2024rq,
  title={Rq-rag: Learning to refine queries for retrieval augmented generation},
  author={Chan, Chi-Min and Xu, Chunpu and Yuan, Ruibin and Luo, Hongyin and Xue, Wei and Guo, Yike and Fu, Jie},
  journal={arXiv preprint arXiv:2404.00610},
  year={2024}
}

@inproceedings{nogueira2017task,
  title={Task-oriented query reformulation with reinforcement learning},
  author={Nogueira, Rodrigo and Cho, Kyunghyun},
  booktitle={Proceedings of the 2017 conference on empirical methods in natural language processing},
  pages={574--583},
  year={2017}
}

@inproceedings{asai2024selfrag,
  title={Self-rag: Learning to retrieve, generate, and critique through self-reflection},
  author={Asai, Akari and Wu, Zeqiu and Wang, Yizhong and Sil, Avirup and Hajishirzi, Hannaneh},
  booktitle={The Twelfth International Conference on Learning Representations},
  year={2024}
}

@inproceedings{mallen2023not,
  title={When not to trust language models: Investigating effectiveness of parametric and non-parametric memories},
  author={Mallen, Alex and Asai, Akari and Zhong, Victor and Das, Rajarshi and Khashabi, Daniel and Hajishirzi, Hannaneh},
  booktitle={Proceedings of the 61st annual meeting of the association for computational linguistics (volume 1: Long papers)},
  pages={9802--9822},
  year={2023}
}

@inproceedings{ke2024bridging,
  title={Bridging the preference gap between retrievers and llms},
  author={Ke, Zixuan and Kong, Weize and Li, Cheng and Zhang, Mingyang and Mei, Qiaozhu and Bendersky, Michael},
  booktitle={Proceedings of the 62nd Annual Meeting of the Association for Computational Linguistics (Volume 1: Long Papers)},
  pages={10438--10451},
  year={2024}
}

@inproceedings{shao2023enhancing,
  title={Enhancing retrieval-augmented large language models with iterative retrieval-generation synergy},
  author={Shao, Zhihong and Gong, Yeyun and Shen, Yelong and Huang, Minlie and Duan, Nan and Chen, Weizhu},
  booktitle={Findings of the Association for Computational Linguistics: EMNLP 2023},
  pages={9248--9274},
  year={2023}
}

@article{GRPO,
  author       = {Zhihong Shao and
                  Peiyi Wang and
                  Qihao Zhu and
                  Runxin Xu and
                  Junxiao Song and
                  Mingchuan Zhang and
                  Y. K. Li and
                  Y. Wu and
                  Daya Guo},
  title        = {DeepSeekMath: Pushing the Limits of Mathematical Reasoning in Open
                  Language Models},
  journal      = {CoRR},
  volume       = {abs/2402.03300},
  year         = {2024},
  url          = {https://doi.org/10.48550/arXiv.2402.03300},
  doi          = {10.48550/ARXIV.2402.03300},
  eprinttype    = {arXiv},
  eprint       = {2402.03300},
  bibsource    = {dblp computer science bibliography, https://dblp.org}
}

@article{thakur2021beir,
  title={Beir: A heterogenous benchmark for zero-shot evaluation of information retrieval models},
  author={Thakur, Nandan and Reimers, Nils and R{\"u}ckl{\'e}, Andreas and Srivastava, Abhishek and Gurevych, Iryna},
  journal={arXiv preprint arXiv:2104.08663},
  year={2021}
}

@article{xiao2025scaling,
  title={Scaling Language-Centric Omnimodal Representation Learning},
  author={Xiao, Chenghao and Chan, Hou Pong and Zhang, Hao and Xu, Weiwen and Aljunied, Mahani and Rong, Yu},
  journal={arXiv preprint arXiv:2510.11693},
  year={2025}
}

@inproceedings{chia2025m,
  title={M-longdoc: A benchmark for multimodal super-long document understanding and a retrieval-aware tuning framework},
  author={Chia, Yew Ken and Cheng, Liying and Chan, Hou Pong and Song, Maojia and Liu, Chaoqun and Aljunied, Mahani and Poria, Soujanya and Bing, Lidong},
  booktitle={Proceedings of the 2025 Conference on Empirical Methods in Natural Language Processing},
  pages={9244--9261},
  year={2025}
}

@article{jin2025search,
  title={Search-r1: Training llms to reason and leverage search engines with reinforcement learning},
  author={Jin, Bowen and Zeng, Hansi and Yue, Zhenrui and Yoon, Jinsung and Arik, Sercan and Wang, Dong and Zamani, Hamed and Han, Jiawei},
  journal={arXiv preprint arXiv:2503.09516},
  year={2025}
}

@inproceedings{xiao2023length,
  title={Length is a curse and a blessing for document-level semantics},
  author={Xiao, Chenghao and Li, Yizhi and Hudson, G and Lin, Chenghua and Al Moubayed, Noura},
  booktitle={Proceedings of the 2023 Conference on Empirical Methods in Natural Language Processing},
  pages={1385--1396},
  year={2023}
}

@article{lewis2020retrieval,
  title={Retrieval-augmented generation for knowledge-intensive nlp tasks},
  author={Lewis, Patrick and Perez, Ethan and Piktus, Aleksandra and Petroni, Fabio and Karpukhin, Vladimir and Goyal, Naman and K{\"u}ttler, Heinrich and Lewis, Mike and Yih, Wen-tau and Rockt{\"a}schel, Tim and others},
  journal={Advances in neural information processing systems},
  volume={33},
  pages={9459--9474},
  year={2020}
}

@inproceedings{xiao2025analyzing,
  title={Analyzing llms’ knowledge boundary cognition across languages through the lens of internal representations},
  author={Xiao, Chenghao and Chan, Hou Pong and Zhang, Hao and Aljunied, Mahani and Bing, Lidong and Al Moubayed, Noura and Rong, Yu},
  booktitle={Proceedings of the 63rd Annual Meeting of the Association for Computational Linguistics (Volume 1: Long Papers)},
  pages={24099--24115},
  year={2025}
}

@inproceedings{xiao2025mieb,
  title={Mieb: Massive image embedding benchmark},
  author={Xiao, Chenghao and Chung, Isaac and Kerboua, Imene and Stirling, Jamie and Zhang, Xin and Kardos, M{\'a}rton and Solomatin, Roman and Al Moubayed, Noura and Enevoldsen, Kenneth and Muennighoff, Niklas},
  booktitle={Proceedings of the IEEE/CVF International Conference on Computer Vision},
  pages={22187--22198},
  year={2025}
}

\newpage
\appendix
\label{sec:appendix}

\section{RAGBench Details}

RAGBench \cite{friel2024ragbench} is a large-scale RAG benchmark dataset which comprises more than 100,000 samples across diverse domains such as finance, law, and healthcare. Specifically, RAGBench collects extensive QA datasets, including CovidQA~\cite{moller2020covid}, PubMedQA~\cite{jin2019pubmedqa}, HotpotQA~\cite{yang2018hotpotqa}, MS Marco~\cite{bajaj2016ms}, CUAD~\cite{hendrycks2021cuad}, EManual~\cite{nandy-etal-2021-question-answering}, TechQA~\cite{castelli2020techqa}, FinQA~\cite{chen2021finqa}, TAT-QA~\cite{zhu2021tat}, ExpertQA~\cite{malaviya2024expertqa}, and HAGRID~\cite{kamalloo2023hagrid}.

\begin{table*}[h]
    \centering
    \renewcommand{\arraystretch}{1.2}
    \resizebox{0.85\linewidth}{!}{\begin{tabular}{l c c c c}
        \toprule
        \textbf{Method} & \textbf{Retrieval (nDCG@10)} & \textbf{Accuracy} & \textbf{F1} & \textbf{EM} \\
        \midrule
        
        \multicolumn{5}{l}{\textbf{Full RAG (w/ top 20 docs)}} \\
        no RAG & - & 25.3 & 26.0 & 18.8 \\
        + vanilla RAG & 41.01 & 41.4 & 41.0 & 31.2 \\
        + general rewrite w/ Gemini & 39.16 & 42.3 & 42.0 & 32.0 \\
        + \textbf{(ours)} retriever-specific rewrite & \textbf{52.44} & \textbf{46.9} & \textbf{46.0} & \textbf{35.2} \\
        
        \midrule
        
        \multicolumn{5}{l}{\textbf{Full RAG (w/ top 5 docs)}} \\
        no RAG & - & 25.3 & 26.0 & 18.8 \\
        + vanilla RAG & 41.01 & 36.5 & 36.0 & 27.3 \\
        + general rewrite w/ Gemini & 39.16 & 38.1 & 38.0 & 28.5 \\
        + \textbf{(ours)} retriever-specific rewrite & \textbf{52.44} & \textbf{43.7} & \textbf{43.0} & \textbf{32.8} \\
        
        \bottomrule
    \end{tabular}}
    \caption{Performance comparison of Full RAG setups using top 20 and top 5 retrieved documents.}
    \label{tab:rag_performance}
\end{table*}

\section{FinAgentBench and its Adaptation}
\label{apx: FinAgentBench}

FinAgentBench \cite{choi2025finagentbench} is a financial information retrieval benchmark which contains over 18,000 samples curated by experts from SEC filings between 2023 and 2024. In the original benchmark, each sample is formulated as a single large prompt that combines ranking instructions, a financial question, and a set of candidate document chunks, which are annotated with graded relevance labels. To adapt FinAgentBench to a standard retrieval setting, we decompose each prompt into its constituent components and extract unique questions and document chunks across all samples. Because the same question may appear in multiple samples (e.g., when posed over different filings), and the same set of document chunks may be reused across samples (e.g., when multiple questions are asked about the same filing), we pool questions and chunks globally and retain only unique instances via exact matching, without any additional normalization or model-based processing. We then inherit relevance annotations from the original benchmark by mapping each question-chunk pair to its corresponding relevance label, thereby constructing a conventional retrieval dataset consisting questions, documents, and graded relevance judgments. 

\section{RE-MMD Empirical Estimation}
\label{app:remmd_details}

In Section \ref{sec:remmd}, we defined RE-MMD as the distance between mean embeddings in a high-dimensional feature space. Directly computing the embedding $\phi(q)$ is often intractable. Therefore, we apply the kernel trick $k(q, q') = \langle \phi(q), \phi(q') \rangle_{\mathcal{H}}$ to estimate the distance empirically.

Given a set of successful query trajectories $Q_{src} = \{x_1, ..., x_n\}$ from $\mathcal{E}_{src}$ and $Q_{tgt} = \{y_1, ..., y_m\}$ from $\mathcal{E}_{tgt}$, we employ the unbiased estimator:

\begin{equation}
\begin{split}
RE\text{-}MMD^2 &= \frac{1}{n(n-1)} \sum_{i \neq j} k(x_i, x_j) \\
&+ \frac{1}{m(m-1)} \sum_{i \neq j} k(y_i, y_j) \\
&- \frac{2}{nm} \sum_{i,j} k(x_i, y_j)
\end{split}
\label{eq:remmd_estimator}
\end{equation}

To diagnose the nature of the drift, we utilize two distinct kernel functions:

\paragraph{1. Semantic Kernel.} To measure intent drift, we use the Gaussian RBF kernel over dense embeddings (e.g., MiniLM):
\begin{equation}
k_{sem}(x, y) = \exp\left(-\frac{||\phi_{sem}(x) - \phi_{sem}(y)||^2}{2\sigma^2}\right)
\end{equation}
This captures non-linear semantic similarities between queries.

\paragraph{2. Structural Kernel.} To measure lexical/stylistic drift, we use a Linear kernel over sparse lexical features (TF-IDF):
\begin{equation}
k_{struct}(x, y) = \langle \phi_{struct}(x), \phi_{struct}(y) \rangle
\end{equation}
This is equivalent to the Euclidean distance of centroids in the lexical space, capturing explicit divergence in vocabulary usage and query formulation style.

\section{Full RAG Performance}
\label{apx:RAG_performance}

To validate whether the performance gains observed in retriever-specific retrieval translate into improvements in the full-RAG pipeline, we conduct a series of full-RAG experiments using HotpotQA as an example. Specifically, we adopt contriever as the retriever and Qwen2.5-7B-Instruct as the generator, and compare our rewriter agent against vanilla RAG and general rewrites (non-retriever-specific) from Gemini-2.5-flash. We use the HotpotQA corpus from BEIR as the external knowledge base and the original HotpotQA questions as queries. Following standard RAG evaluation protocols, we report Accuracy, F1, and Exact Match (EM).

As shown in Table~\ref{tab:rag_performance}, enhancements in nDCG@10 consistently translates to performance gains in the full RAG performance. Notably, these improvements remain stable across different numbers of retrieved documents, indicating the robustness of retriever-aware query rewriting.

\section{Zero-shot Multi-turn Retrieval Case Study}
\label{apx:multi-turn case study}

We provide an illustrative example of multi-turn retrieval using our rewrite agent over four retrieval turns. Although the model is trained using only a two-turn branching strategy in our experiments, it demonstrates the ability to generalize to longer multi-turn interactions in a zero-shot manner. This suggests that the learned retrieval behaviors are not limited to the training horizon and can effectively extend to more complex interaction settings.

\vspace{2mm}
\noindent \textbf{Original Query:} \textit{"What would be the effects on the us dollar if hypothetically it is no longer regarded as the world reserve currency?"}
\vspace{2mm}

\noindent\fbox{%
    \begin{minipage}{0.96\linewidth}
    \small
    \textbf{Rewrite Turn 1 (Hypothetical Document Generation):} \\
    The US dollar's status as the world's primary reserve currency has been a cornerstone of global economic stability for decades. If the dollar were hypothetically no longer regarded as the world reserve currency, the implications would be profound and multifaceted. First, the international demand for the dollar would likely decline, leading to a depreciation in its value. This depreciation could increase the cost of imports for the United States, contributing to inflation. Conversely, US exports might become more competitive, potentially boosting the economy in the short term.
    
    \vspace{1mm}
    Central banks and governments around the world currently hold significant reserves of US dollars, using them for trade settlements, investments, and as a safe asset. A shift away from the dollar would require these entities to diversify their reserves, possibly increasing the role of other currencies such as the euro, the Chinese yuan, or the Japanese yen. This transition could lead to the emergence of new global financial centers and alter the balance of economic power.
    
    \vspace{1mm}
    The United States would also face challenges in maintaining its influence over global financial institutions, such as the International Monetary Fund (IMF) and the World Bank, which have historically been dominated by US interests. A reduced role of the dollar could diminish the US's ability to shape international economic policy, affecting everything from trade agreements to financial regulations.
    
    \vspace{1mm}
    Moreover, the US government's capacity to borrow at low interest rates might be impacted, as the dollar's status as a safe haven could diminish. This could lead to higher borrowing costs for the US, affecting both the federal budget and private sector investments. In the long term, the global economy might become more fragmented, with regional currencies playing a larger role in international trade, potentially leading to increased economic volatility.
    
    \vspace{1mm}
    While the transition away from the dollar as the reserve currency would present challenges, it could also offer opportunities for the US to innovate in financial technology and strengthen its economic partnerships with other nations. The ultimate impact would depend on how smoothly the transition is managed and the policies implemented to support a new global financial order.
    \end{minipage}%
}

\vspace{1.5mm}
\noindent \textbf{Analysis:} The rewriter uses the \textbf{hypothetical document generation} technique in turn 1, but the terms are too general, yielding a \textbf{nDCG@10 of 0.0} in the first round. However, the rewriter is able to see the returned documents and learns how documents in the corpus look like.

\vspace{4mm}

\noindent\fbox{%
    \begin{minipage}{0.96\linewidth}
    \small
    \textbf{Rewrite Turn 2 (Entity Anchoring):} \\
    What are the economic implications of \textbf{de-dollarization} and the rise of alternative reserve currencies, such as the \textbf{Chinese yuan}, on \textbf{global financial systems} and the United States?
    \end{minipage}%
}

\vspace{1.5mm}
\noindent \textbf{Analysis:} Feedback from turn 1, although low scoring, exposes specific entities in the returned documents (omitted here). Thus, in the second round, the rewriter pivots to more specific directions, such as "de-dollarization", "Chinese yuan", and "global financial systems", enhancing the \textbf{nDCG@10 to 56.84}.

\vspace{4mm}

\noindent\fbox{%
    \begin{minipage}{0.96\linewidth}
    \small
    \textbf{Rewrite Turn 3 (Further Enhancement):} \\
    \textit{\{omitted for brevity\}}
    \end{minipage}%
}

\vspace{4mm}

\noindent\fbox{%
    \begin{minipage}{0.96\linewidth}
    \small
    \textbf{Rewrite Turn 4 (Domain/Style Alignment):} \\
    What are the financial and economic ramifications for the United States and global markets if the US dollar were no longer the dominant reserve currency, including impacts on foreign exchange reserves, \textbf{corporate financial strategies}, and the role of alternative currencies like the Chinese yuan?
    \end{minipage}%
}

\vspace{1.5mm}
\noindent \textbf{Analysis:} After understanding the genre of available documents (corporate 10-K filings and financial disclosures) in this corpus, the agent adopts specific financial reporting terminology and be more specific about the domain, adapting to the environment style. The final rewrite enhances the \textbf{nDCG@10 to 93.74}, a near perfect score.

\section{Training Prompts}
\label{apx: training prompts}
Figure~\ref{fig:training_prompts} describes the training prompts used in different settings of our experiments.

\begin{figure*}[t!]
    \centering
    \vspace{-5mm}
    \fbox{
        \begin{minipage}{\linewidth}
            \small
            \textbf{General Prompt:} \\
            In order to retrieve supporting documents to answer the given question, you need to rewrite and extend a given question into one new query. Remember that the rewritten query is not limited to a simple question, but can be a passage including multiple sentences that provides context for the input question. Here are some commonly used query rewriting methods, please select one appropriate method and use it to generate one new rewritten query: \\ 1. Query Understanding: Extract core information from the query, and understand the core elements. For example, `What was the federal alternative minimum tax credit carryforwards?' can be phased as `Federal, U.S. federal tax system; Alternative Minimum Tax (AMT), a parallel tax system designed to ensure taxpayers pay a minimum level of tax; Credit carryforwards, unused credits that can be carried forward to future years to offset regular tax liability;'. \\2. Semantic Rewriting: (1) Identify important entities and expand with related information. For example, `Steve Jobs' to `Apple CEO Steve Jobs'. (2) Add broader (hypernym) or more specific (hyponym) concepts. For example, `husky' to `pet dog husky'. (3) Change word order or phrasing to match common expressions in target documents. For example, `make steak' to `steak recipe tutorial'. \\3. Query Decomposition: Break down a complex query into multiple sub-queries for separate retrieval. For example, `Which stock price is higher, Apple or Google?' to `<rewrite>what is the stock price of Apple?</rewrite><rewrite>what is the stock price of Google?</rewrite>'. Also consider more complex queries that can benefit from breaking down into multi-hop queries. \\4. General Rewriting: (1) Paraphrasing: rephrase the whole query if you think can possibly help the embedding model generate more accurate embeddings. For example, `How to lose weight effectively?' to `What are some scientifically proven weight loss methods?'. (2) Intent Refinement, infer the user true intent and rewrite the query accordingly. (3) Multi-hop Query Generation, generate indirectly related queries that help improve recall. For example, `diabetic diet' to `fruits suitable for diabetics'. \\5. Hypothetical Document Generation: Generate a hypothetical target document, in order to use this hypothetical document to retrieve target document. For example, (1) Given a question asking about the law clause related to a case, directly generate a hypothetical clause for retrieval. (2) Given a question about a company's stock, directly generate a hypothetical stock price report that looks alike the target document that contains the information. \\ Before outputting the rewritten query, you need to think carefully about choosing which rewriting method, why choose this method, and how to use it. And please enclose your think process within <think> and </think>. Then the rewritten query must be enclosed between <rewrite> and </rewrite>. Do not put the thinking process between <rewrite> and </rewrite>. For methods other than Query Decomposition, only one rewrite query is needed. If the question is rewritten as mutiple queries using one rewriting method (for example, when using Query Decomposition), please enclose each different query with separate <rewrite> and </rewrite>. Output can look like: <think> your think process </think><rewrite> rewritten query </rewrite>. User Question:  
            \vspace{2mm}
            \hrule
            \vspace{2mm}
            
            \textbf{Exploratory Prompt (Contriever):} \\
            You are using a retriever model to retrieve relevant documents to answer a question. The retriever model you will be using is Contriever, which is trained using two spans of a Wikipedia article as positive pairs. You will need to think about the best search strategy using Contriever and rewrite the original query into a query that you think yields the best retrieval results for Contriever. Before outputting the rewritten query, you need to think carefully about what to write to look like target documents, enclose your think process within <think> and </think>. Then the rewritten query must be enclosed between <rewrite> and </rewrite>. Do not put the thinking process between <rewrite> and </rewrite>. Output can look like: <think> your think process </think><rewrite> rewritten query </rewrite>. 
User Question:
            
            \vspace{2mm}
            \hrule
            \vspace{2mm}
            
            \textbf{Exploratory Prompt (bm25)} \\
            You are using a retriever system to retrieve relevant documents to answer a question. The retriever system you will be using is BM25, which is a sparse retrieval system. You will need to think about the best search strategy using BM25 (e.g., keyword-based) and rewrite the original query into a query that you think yields the best retrieval results for BM25. Before outputting the rewritten query, you need to think carefully about what to write to yield the best retrieval results, enclose your think process within <think> and </think>. Then the rewritten query must be enclosed between <rewrite> and </rewrite>. Do not put the thinking process between <rewrite> and </rewrite>. Output can look like: <think> your think process </think><rewrite> rewritten query </rewrite>. 
User Question:
            \vspace{2mm}
            \hrule
            \vspace{2mm}
            \textbf{Direct/Specific Prompt (Contriever)} \\
            You are using a retriever model to retrieve relevant documents to answer a question. The retriever model you will be using is Contriever, which is trained on using two spans of a Wikipedia article as positive pairs. Therefore, it works best for using document to retrieve document. 
You will be given a user question, try using the "Hypothetical Document Generation" technique to generate a hypothetical target document, which will be used to retrieve target similar documents. For example, (1) Given a question asking about the law clause related to a case, directly generate a hypothetical clause for retrieval. (2) Given a question about a company's stock, directly generate a hypothetical stock price report that looks alike the target document that contains the information. 
Before outputting the hypothetical document, you need to think carefully about what to write to look like target documents, enclose your think process within <think> and </think>. Then the hypothetical document must be enclosed between <rewrite> and </rewrite>. Do not put the thinking process between <rewrite> and </rewrite>. Output can look like: <think> your think process </think><rewrite> hypothetical document </rewrite>. 
User Question:
            \vspace{2mm}
            \hrule
            \vspace{2mm}
            \textbf{Direct/Specific Prompt (bm25)} \\
            You are a query optimization assistant for a BM25 retrieval system. Since BM25 is a sparse, keyword-based model, it performs best with precise search terms rather than natural language. Your task is to rewrite the user's original query into a keyword-optimized version that maximizes retrieval performance. You must specifically optimize for term morphology and spelling variations (such as singular vs. plural forms or compound word handling). Because BM25 relies on exact token matching to calculate Inverse Document Frequency (IDF), using the most standard or canonical form of a word is critical for aligning with the target corpus. Output the best possible search query for this system. Before outputting the rewritten query, you need to think carefully about what to write to yield the best retrieval results, enclose your think process within <think> and </think>. Then the rewritten query must be enclosed between <rewrite> and </rewrite>. Do not put the thinking process between <rewrite> and </rewrite>. Output can look like: <think> your think process </think><rewrite> rewritten query </rewrite>. 
User Question:
        \end{minipage}
    }
    \caption{Training prompts used in our different settings.}
    \label{fig:training_prompts}
\end{figure*}
\end{document}